%% file: main.tex
\documentclass[]{fairmeta}

\usepackage{graphicx} 
\usepackage{amsmath,amsfonts,amsthm}
\usepackage{fullpage}
\usepackage{color}
\usepackage{booktabs}
\usepackage{multirow}
\usepackage{tabularx}
\usepackage{url}

\newcommand{\had}{\odot}
\newcommand{\argmin}{\mathbf{argmin}}

\newtheorem{proposition}{Proposition}

\input{macros}

\title{Privacy Blur: Quantifying Privacy and Utility for Image Data Release}

\author[1,*]{Saeed Mahloujifar}
\author[1,*]{Narine Kokhlikyan} 
\author[1]{Chuan Guo}
\author[1]{Kamalika Chaudhuri}

\affiliation[1]{FAIR at Meta}
\contribution[*]{Joint first author}

\abstract{Image data collected in the wild often contains private information such as faces and license plates, and responsible data release must ensure that this information stays hidden. At the same time, released data should retain its usefulness for model-training. The standard method for private information obfuscation in images is Gaussian blurring. In this work, we show that practical implementations of Gaussian blurring are reversible enough to break privacy. We then take a closer look at the privacy-utility tradeoffs offered by three other obfuscation algorithms -- pixelization, pixelization and noise addition (DP-Pix), and cropping. Privacy is evaluated by reversal and discrimination attacks, while utility by the quality of the learnt representations when the model is trained on data with obfuscated faces. We show that the most popular industry-standard method, Gaussian blur is the least private of the four -- being susceptible to reversal attacks in its practical low-precision implementations. In contrast, pixelization and pixelization plus noise addition, when used at the right level of granularity, offer both privacy and utility for a number of computer vision tasks. We make our proposed methods together with suggested parameters available in a software package called Privacy Blur. }

\begin{document}

\maketitle

\input{intro.tex}

\input{relwork.tex}

\input{prelim.tex}

\input{attack.tex}
\input{theory_new.tex}

\input{experiments}

\section{Conclusion}

In conclusion, we investigate the privacy-utility tradeoffs of different methods for obfuscating private information in images. We find that the industry standard method, Gaussian blurring, is the least private and easily reversible. In contrast, options like pixelization and DP-Pix, when used with the right parameters, provide a nice mix of both privacy and utility on downstream model training. We make our proposed methods together with suggested parameters available in a software package called Privacy Blur. We hope that our investigation will inform and enable more privacy-aware methods for future image data releases. 

\bibliographystyle{plainnat}
\bibliography{blur}

\section*{Acknowledgements}
This paper has benefitted from advice and discussion from many people. We thank Mike Rabbat for many helpful suggestions and advice on framing this work, and Joelle Pineau, Mary Williamson and Kristin Lauter for their continued support and championship. We thank Carolyn Krol for sharing her insights on privacy and risk, and Elisa Cascardi and Philippe Brunet for being instrumental in enabling this work.

\end{document}

%% file: macros.tex
\usepackage{amsmath}
\usepackage{amssymb}
\usepackage{amsthm}
\usepackage{appendix}
\usepackage{bbm}
\usepackage{mathtools}
\newcounter{thm}
\newtheorem{definition}[thm]{Definition}

%% file: intro.tex
\section{Introduction}

Data is the new oil that will power the AI engines of tomorrow. Fueled by the need to train bigger and better computer vision models, the last few years have seen a significant rise in the number, size and quality of released image datasets. At the same time, many of these datasets, especially those collected in the wild, contain private information, such as, faces, house numbers and license plates. Responsible data release must ensure that this information stays hidden, while still retaining the usefulness of the released data for model-training.

The typical pipeline for data release has two steps. First, regions of the image containing private information such as a face or a license plates are detected using an automatic detector; these regions are then obfuscated. The standard way for obfuscation is Gaussian blurring: this convolves the image with a Gaussian kernel and renders it visually unrecognizable. Practical forms of Gaussian blurring are implemented by libraries such as PIL and OpenCV, and sometimes involve multiple iterations of blurring together with rounding to lower precision for efficiency. 

In this work, we take a closer look at this second step of obfuscation and measure its privacy-utility tradeoff in detail. Prior work~\cite{yang2022study} has looked at the evaluating only the accuracy of models when trained on Imagenet where faces are Gaussian blurred or cropped. In contrast, we look at a broader spectrum of obfuscation methods, datasets and tasks, and evaluate {\em{both}} privacy and utility. In particular, in addition to Gaussian blurring and cropping, we look at two other obfuscation methods: pixelization, which involves dividing the image into blocks and replacing each block by the average pixel color, and pixelization followed by noise-addition (DP-Pix~\cite{fan2019differential}). We also look at two more relevant computer vision tasks: representation learning on MetaCLIP and pose estimation on MS-COCO.

We evaluate privacy via two different empirical attacks. The first is the reversal attack, where the goal is to partially or completely reconstruct the private information present in an obfuscated image; for example, reconstruct an entire face so that it is recognizable. We introduce a new optimization-based method for executing a reversal attack. A second, weaker, attackis the discrimination attack, which has also been considered in prior work~\cite{mcpherson2016defeating, hill2016effectiveness}. Here the private information can only take a limited number of options (such as, letters and digits), and our goal is to determine the option the obfuscated information came from, given an auxiliary sample of obfuscated images and their labels. For example, if we are looking to obfuscate an image which could be either $0$ or $1$, then our goal is to infer which digit it is, given a sample of similarly obfuscated $0$s and $1$s. 

Our experiments reveal an interesting pattern. In terms of privacy, as expected, cropping is the most private and is not susceptible to either attack. However, Gaussian blurring, the most popular obfuscation method is highly non-private in its practical implementations, and is susceptible to both reversal and discrimination attacks. In contrast, pixelization and DP-Pix, when used with the right parameters, cannot be reversed reliably. In addition, in simple OCR datasets such as MNIST these methods are amenable to discrimination attacks, but in more complex datasets such as SVHN, they are less so. 

In terms of utility, we find that cropping, Gaussian blurring and pixelization and DP-pix form a privacy-utility curve. Cropping is the most private and the least accurate, particularly in use-cases involving pose estimation. Gaussian blurring is the least private and most accurate. Pixelization and DP-Pix fall somewhere in between the two extremes - more privacy than Gaussian blurring for slightly less accuracy. We also find that for most tasks, the loss of utility of these two methods over Gaussian blurring is quite minimal. Our findings suggest that pixelization and DP-Pix are good alternatives to Gaussian blurring for private information obfuscation in images. 

To summarize, our contributions are as follows:
\begin{itemize}
\item We propose a new privacy measurement framework for private information obfuscation in image with two empirical tests: reversal and discrimination attacks.
\item We provide a simple and novel reversal attack that applies to practical versions of Gaussian blurring. 
\item We provide a novel privacy analysis for different image obfuscation methods. 
\item We evaluate four image obfuscation methods for privacy and utility. We find that Gaussian blurring is the least private, being susceptible to reversal and discrimination attacks, and less private pixelization and DP-Pix.
\item We find that Gaussian blurring has the highest utility, followed by pixelization and DP-Pix, followed by cropping. While cropping has a bit lower utility for pose estimation tasks, for most standard benchmark tasks, the drop in accuracy due to face obfuscation is minimal.
\end{itemize}

%% file: relwork.tex
\section{Related Work}

There is a body of scientific literature on privacy in visual data as well as privacy-preserving machine learning.

\subsection{Obfuscation Methods} 

The standard approach to releasing image data while ensuring privacy is to detect sensitive information such as faces and license plates and then obfuscate them. The main methods available for obfuscation are blurring, pixelization (also known as mosaicing), noise-addition (such as DP-Pix~\cite{fan2019differential}), and in-painting~\cite{yu2018generative}. 

The most popular method is blurring, where private information is convolved with a Gaussian kernel. This leads to good visual quality, but private information in the form of pixels still remains in the image even though not in human-perceptible form. Examples include \cite{frome2009large}, which detects and blurs faces and license plates in Google Street View, and recently released image and video datasets such as the AViD dataset~\cite{piergiovanni2020avid} among others. 

An alternative to blurring is pixelization, also known as mosaicing, where the image is broken down into blocks, and the average value of pixels in each block is reported. This offers more privacy due to information loss, but is used less frequently in data release due to poor visual quality of the resulting images. 

Motivated by metric differential privacy, \cite{fan2019differential} proposes methods called DP-Pix and its variation DP-Blur, which essentially involve pixelization, followed by noise addition, followed by Gaussian blurring. While this offers more privacy than plain pixelization, it has not been widely used in data release, possibly because its impact on downstream representation learning is unknown. 

Finally, perhaps the most private obfuscation method is cropping, which replaces private information by a black box; it is also the most visually jarring. Related to cropping is in-painting, which replaces private information with an AI-generated image. In-painting however is highly dependent on the quality and details of the in-painting model, and may lead to issues with scaling if this model is too slow, or privacy issues depending on the training set of the in-painting model. 

\subsection{Privacy Attacks on Blurring and Pixelization}

In general, blurring has no privacy guarantees, and retains almost all information about facial features in the image. Hence, it can be reversed -- in fact, reversing high-precision blurring has been well-studied as deconvolution! It is worth noting that practical implementations of blurring are rarely full-precision. Industry-standard blurring employs practical libraries such as OpenCV or PIL, which often combine multiple iterations of blurring together with rounding to lower precision for efficiency. We provide a new reversal attack that shows that even these methods can be approximately reversed, and may not be secure enough for private data release. Note however that unlike blurring, methods such as pixelization and DP-Pix, when used with the right parameters, cannot be reversed since they involve information loss. The same holds for cropping. 

Separate from reversal, prior work has also looked at discrimination or recognition attacks on obfuscated images for letters and digits: given an obfuscated image of a digit, the goal is to distinguish it from the remaining nine given auxiliary samples of similarly obfuscated digits and their labels. \cite{hill2016effectiveness} shows that letters and numbers from printed documents can be recognized from their pixelized versions using a Hidden Markov Model-based classifier; most of their results are on finer-grained pixelization (than what we use). A similar result that uses neural networks is shown in~\cite{mcpherson2016defeating}. However, the scope of both results are somewhat limited: datasets are small-scale and somewhat toy,  recognition tasks only have a small handful of options, and pixelization is quite fine-grained. Our experiments show that while similar attacks succeed on blurring, they are not as effective on pixelization and DP-Blur when used with the right parameters on more realistic datasets of digits and license plates. 

\subsection{Utility} 

The main goal of image data release is to enable its use in training models for tasks such as representation learning and object recognition. \cite{yang2022study} is the first to show that data where faces are obfuscated by Gaussian blurring is up to this task, and models trained on face-blurred Imagenet can learn high quality representations. \cite{othercvprpaper} shows that pose recognition can also succeed on face-blurred images. Our work completes the picture by investigating representation learning on data where faces are obfuscated by pixelization, DP-Pix and cropping. We show that as expected, cropping leads to the worst quality representations, while pixelization and DP-Pix leads to representations of slightly worse quality than cropping.

\subsection{Machine-Learning with Guaranteed Privacy.} 

There is also a body of literature on machine learning with guaranteed privacy~\cite{chaudhuri2011differentially, abadi2016deep}, the main notion of privacy there being differential privacy~\cite{dwork2006calibrating}. This literature mostly applies in a centralized setting where raw private data is held by a trusted party (and not modified), while the trained model is made publicly available. Differential privacy ensures that the participation of a single person in the private dataset does not impact the probability of any outcome by much; this is ensured by training models through DP-SGD~\cite{song2013stochastic, abadi2016deep} that clips gradients and adds noise during training. 

In contrast, our goal is to release the data itself publicly while preserving the privacy of people in the data. It is worth noting that there has been some prior work on differentially private data-release~\cite{dwork2006calibrating, dwork2009complexity, aydore2021differentially}; however, this is generally considered to be a very challenging task. A number of impossibility results~\cite{dwork2006calibrating} show that data release in its full generality is not possible with differential privacy, and solutions with high privacy-utility tradeoffs are available only for fairly restricted settings such as tabular data~\cite{aydore2021differentially, gupta2012iterative}.

%% file: prelim.tex
\section{Preliminaries}

We represent an image by a $c \times w \times h$ matrix where $c$ is the number of color channels, $w$ is the width in pixels and $h$ is the height. We use the notation $\bf{0}$ (resp. $\bf{1}$) to represent the all-zeros (resp. all-ones) matrix, and $\had$ to denote the element-wise product of two matrices. 

Obfuscation of private information in an image works in two steps. First, a detector is used to find the region of an image which contains private information, such as a face or a license plate. This region is represented by a {\em{mask}}. For a $c \times w \times h$ image $X$, a mask is a $w \times h$ zero-one matrix $M$, whose entries are $1$ if the corresponding pixel lies within a face or a license plate and zero otherwise. The second step is to obfuscate the private information; this can be done in multiple ways, leading to different privacy and utility tradeoffs. 

For the purpose of this work, we assume that a good detection algorithm is available, and concern ourselves only with the obfuscation part. 

\subsection{Obfuscation Methods}
\label{sec:obfuscation}

We consider four different obfuscation methods -- cropping, Gaussian blur, pixelation and DP-Pix. 

\paragraph{Cropping.} The simplest, most private method is cropping, which replaces the pixels in the private region by zero. Formally, 
\begin{equation}\label{eq:crop}
\textbf{crop}(X) = X \had ({\mathbf{1}} - M) + {\mathbf{0}} \had M,
\end{equation}
Cropping is extremely private in the sense that no information about the private region is left in the image; it is however quite visually jarring. In addition, as shown by our experiments, representations learnt on data where faces are obfuscated by cropping have the worst quality out of all the methods considered.   

\paragraph{Gaussian Blur.} The most popular obfuscation method is Gaussian blurring, where the entire image is convolved with a Gaussian kernel, and then the private region is replaced by the convolved version. Formally, if $G$ is a Gaussian kernel, then,
\begin{equation}\label{eq:gblur}
\textbf{blur}(X) = X \had ({\mathbf{1}} - M) + (X * G) \had M
\end{equation}
The Gaussian kernel has a parameter kernel-size, which measures the width of the kernel -- higher width means smoother convolution. Observe that the convolution operation computes a weighted average of the pixel values; consequently the private information is not lost, but present in a more diffuse form. Out of the methods considered, Gaussian blur has the best utility - it is the most visually appealing, and leads to the highest quality representations as shown by our experiments. 

\paragraph{Pixelization.} $m \times n$-pixelization breaks up the $w \times h$ image into $m \times n$ blocks; all pixels in a block are then replaced with the average pixel value in the block. The private region of the image is then replaced by the pixelized version. 
\begin{equation}\label{eq:pixel}
\textbf{pixelize}(X) = X \had ({\mathbf{1}} - M) + X_{m \times n} \had M,
\end{equation}
where $X_{m \times n}$ is the $m \times n$-pixelized version of the image. Observe that unlike blurring, pixelization offers privacy through information loss; instead of the details of the facial features for example, we retain only $m \times n$ pixels worth of information. The lower $m$ and $n$ are, better is the privacy guarantee. Visually however, pixelization is not very attractive, and as our experiments show, lead to representations that are better than cropping but worse than Gaussian blur. 

\paragraph{DP-Pix.} DP-Pix~\cite{fan2019differential} pixelizes an image into $m \times n$ blocks, adds independent Gaussian noise to each block, and finally, applies Gaussian blur to the resulting image in order to enhance its visual quality. To obfuscate only private information, we consider a variation of DP-Pix where the private region of the image is replaced by the corresponding DP-Pix version.
\begin{equation}\label{eq:dppix}
\textbf{DP-pix}(X) = X \had ({\mathbf{1}} - M) + (X_{m \times n} + 
 N_{m \times n}) \had M,
\end{equation}
where $N_{m \times n}$ is a $m \times n$ vector of Gaussian noise with standard deviation $\sigma$. DP-Pix offers privacy through both information loss and information obfuscation through randomization, and the level of privacy offered depends on both the standard deviation of noise added as well as the granularity of the pixelization: higher $\sigma$ and lower values of $m$ and $n$ imply higher privacy. \cite{fan2019differential} provided a metric differential privacy guarantee; we, in our work, provide an additional more interpretable guarantee on its performance. In terms of utility, we find that it also depends on the amount of noise added, with more noise leading to worse utility, both in terms of visual quality and in terms of the quality of learnt representations.

\subsection{Privacy}

When we hide private information in images, the amount of privacy offered depends on the granularity and diversity of the private information. For example, it is challenging to ensure that images of zeroes and ones are indistinguishable after obfuscation since there are only two options which look quite different. On the other hand, it is a much easier task to obfuscate a face so that it is indistinguishable from a sea of other faces that may look similar.

To capture this, we consider two kinds of privacy attacks -- reversal and discrimination. 

\paragraph{Reversal Attack.} In a {\em{reversal}} attack, the goal of the adversary is to partially or completely reconstruct the private information present in an obfuscated image. If a reversal attack is possible, then an adversary may be able to recognize someone and distinguish them from others who look somewhat alike; this makes it a strong attack.

\paragraph{Discrimination Attack.} A weaker attack is {\em{discrimination}}, also studied by~\cite{hill2016effectiveness} and~\cite{mcpherson2016defeating}. Given obfuscated versions of private information which can only take a limited number of options (such as, letters and digits), the goal of a discrimination attack is to determine which option the obfuscated image came from. If there are a small number of options and the data is relatively clean, then a discrimination attack may still be possible even though a reversal attack is not. 

Finally, note that in this work, we only confine ourselves with {\em{direct}} de-obfuscation of private information, and not with indirect inference from side information. An example of the latter is an image of a soccer player in uniform with their face cropped out; even though there is no facial information, an adversary may still be able to infer the player's identity from the team uniform and player number. This kind of indirect inference is outside of the scope of this work. 

\subsection{Utility} 

For general-purpose data release, one notion of utility is visual quality of the image after obfuscation. This is however hard to measure quantitatively. 

Many image datasets such as Imagenet however are released to facilitate model training; in those cases, the relevant notion of utility is the quality of the learnt representations. The latter is measured by standard computer vision benchmarks, such as zero-shot accuracy on classification on benchmark datasets.

%% file: attack.tex
\section{A Reversal Attack}
\label{sec:reversal}

We next propose a simple reversal attack and instantiate it for existing blurring operations in standard libraries. Our attack applies to a white-box adversary, who has access to the exact obfuscation algorithm and its detailed implementation (including, padding and quantization operations) and any associated parameters. If part of the image is obfuscated, then the adversary also knows the {\em{mask}} that describes the obfuscation region, as well as the exact operation that combines the blurred and un-blurred regions. 

Note that all of these assumptions are made to isolate the privacy of the blurring operation itself. In other words, we idealize the setting for adversary to evaluate privacy in the worst-case scenario. While this is a strong requirement on an adversary, it is somewhat realistic -- some data releases come with associated obfuscation code, and in some cases, adversaries may be able to guess the masked region by visual inspection and trial-and-error.   

\paragraph{Attack Description.}
Let $x\in X$ denote an image $x$ in the image space $X$. Let $G: X\times Aux \to X$ be a \emph{deterministic} obfuscation operator, applied on an input $x$ and  a set of parameters $p\in Aux$, so that the obfuscated version of an image $x$ is $y = G(x, p)$. The parameter set $p$ includes parameters such as the kernel size and variance of the Gaussian blur operation, or a mask for cases where blurring is performed only in a certain region.

Given $y$ andx $p$, the goal of the reversal adversary is to find an image $x'$ such that $G(x^*,p) \approx G(x,p)$. Formally, 
\begin{equation} \label{eqn:rev}
x^* = \argmin_{x} \| y - G(x,p).\|
\end{equation}
We use the $\ell_2$ distance as a distance metric in our formulation above, but one can replace that with any other distance metric. We also assume that the obfuscation algorithm is deterministic -- for a randomized obfuscation algorithm, the objective would be likelihood maximization which is out of scope for our attack.   

Note that if $G$ is differentiable with respect to the second parameter, then one can attempt at finding a good solution to this optimization problem by performing gradient descent. The rest of this section is devoted to the subtleties of calculating the gradient for popular blurring algorithms.

\paragraph{Gaussian blurring:} 
In its simplest form, Gaussian blurring is a linear transformation that convolves a Gaussian kernel of certain size and standard deviation with the unblurred image.  To reverse it, we can calculate the gradient of the transformation and apply the optimization attack described above. However, there are multiple subtle challenges that makes attacking standard implementations of Gaussian blurring hard -- which is perhaps why previous work has not tried them. Below, we list some of these challenges:

\begin{enumerate}
   
    \item Approximations and Quantization: In order to make the blurring more efficient some Gaussian blurring mechanisms use approximations of true Gaussian blurring that make them harder to attack. These approximations often come in form of quantization or entirely changing the linear operation. The data-dependent nature of these approximations make the attack significantly harder. 
    \item Parameterization: Existing libraries for blurring often use data/mask dependent parameters that are different from the parameterization of Gaussian blurring. Decoding the role of these parameters and their connection to Gaussian convolution is a difficulty in performing an attack. 
\end{enumerate}

We explore two existing libraries that implement Gaussian blurring and try to overcome the challenges above.

\begin{figure}[h]
    \centering
    \begin{minipage}{0.24\textwidth}
        \centering
        \includegraphics[width=\textwidth]{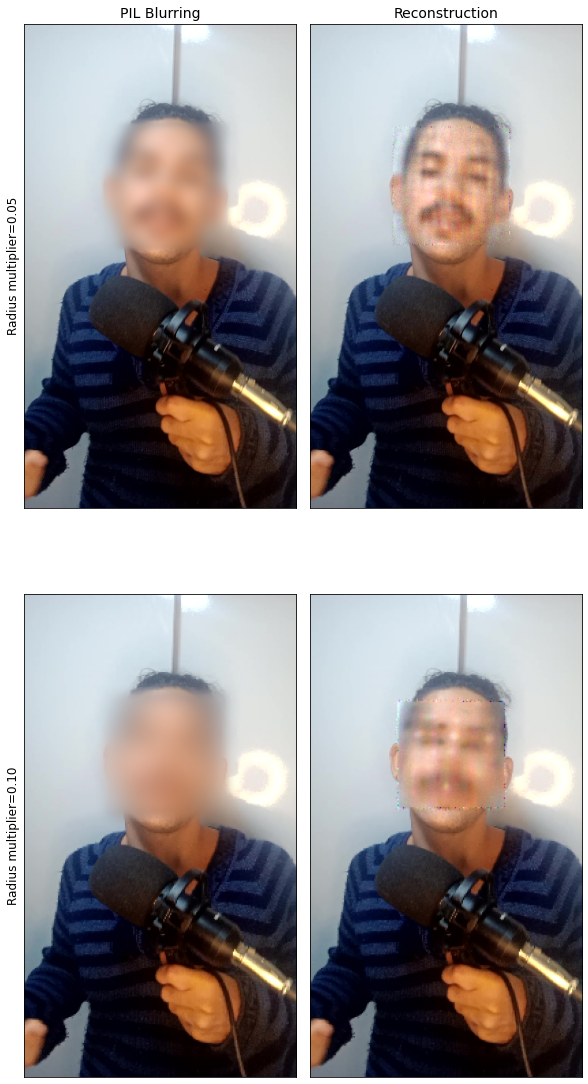}
    \end{minipage}\hfill
    \begin{minipage}{0.24\textwidth}
        \centering
        \includegraphics[width=\textwidth]{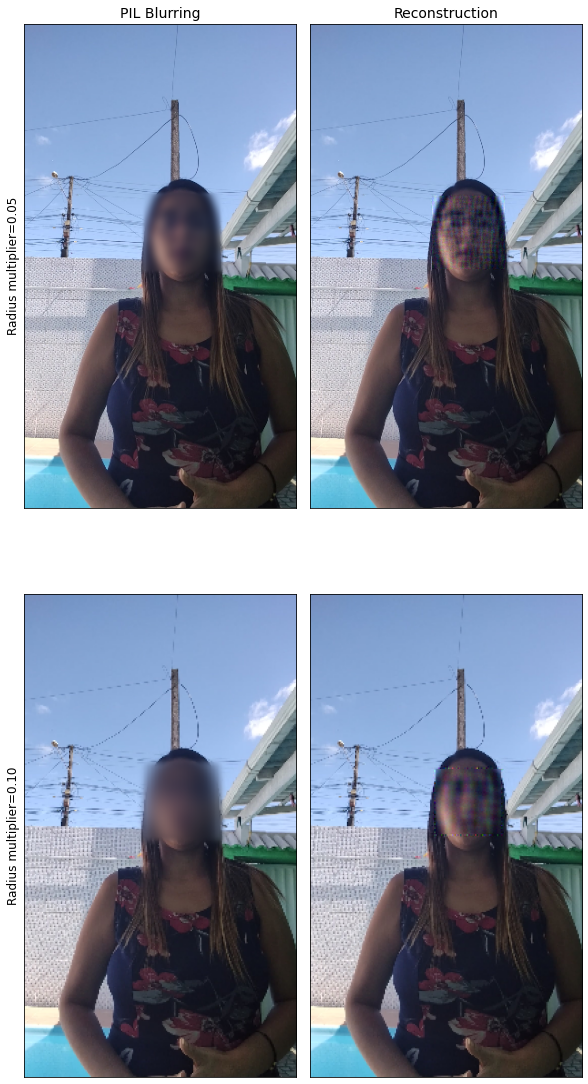}
    \end{minipage}\hfill
    \begin{minipage}{0.24\textwidth}
        \centering
        \includegraphics[width=\textwidth]{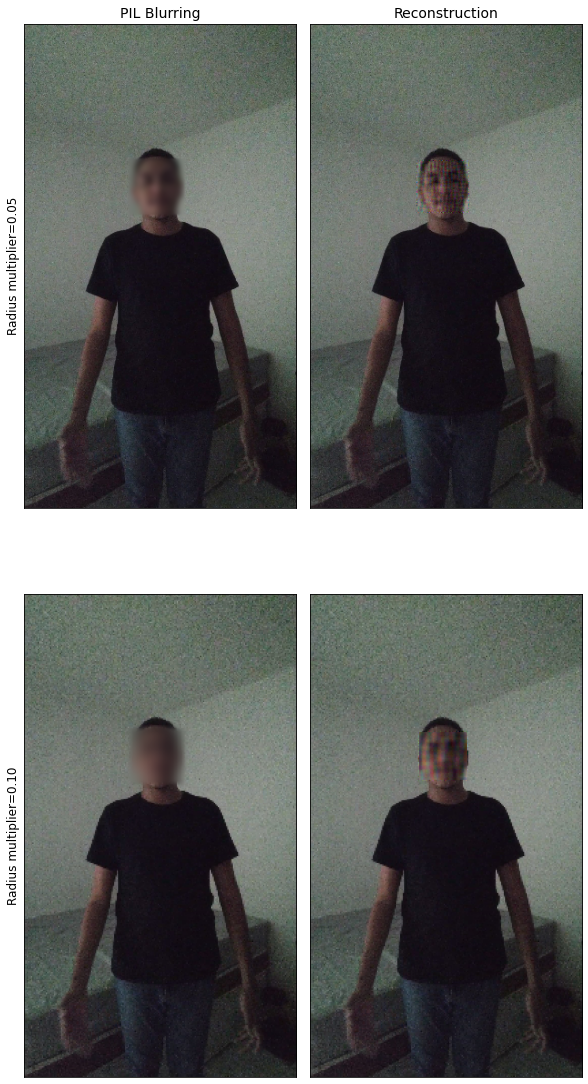}
    \end{minipage}\hfill
    \begin{minipage}{0.24\textwidth}
        \centering
        \includegraphics[width=\textwidth]{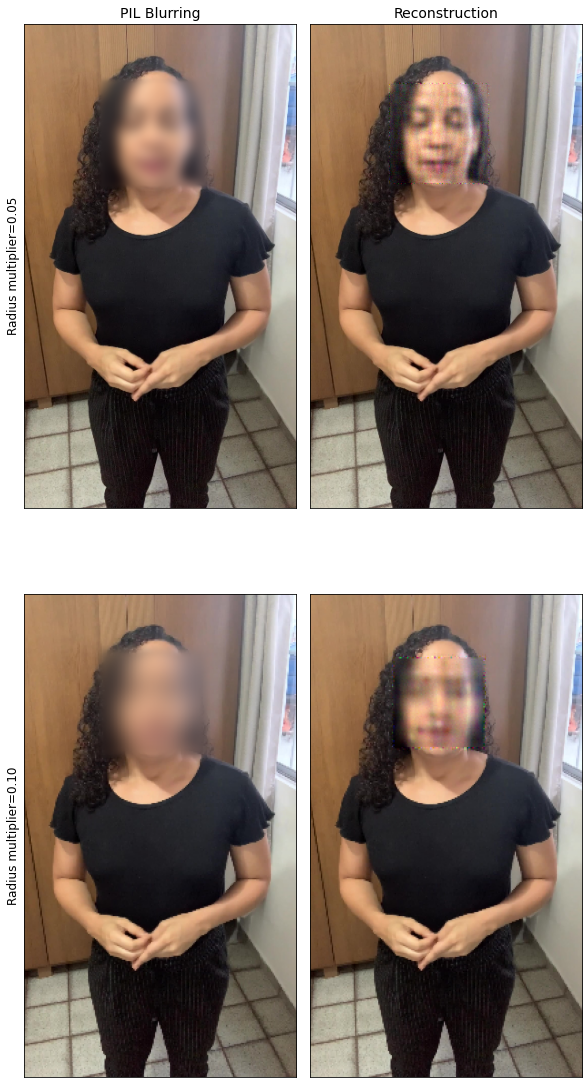}
    \end{minipage}
    \caption{We use random images from  Casual Conversations v2 dataset. We use deep-face to create face bounding boxes and use blurring pipeline in \cite{yang2022study} to blur images. We use two different radius multipliers. Our attack is initialized randomly (Gaussian noise) and uses learning rate 0.1 and momentum of 0.9. We run the attack for 5000 iterations. We then run the attack for 30 times and take the average over all runs. }
    \label{fig:PIL}

\end{figure}

\begin{itemize}
    \item PIL Library: PIL is a widely used image processing library. The Gaussian blurring implementation takes a single parameter called radius and blurs the image using that. Based on the value of radius, the operation is expected to approximate Gaussian blurring with a specific kernel size and standard deviation. However, the exact implementation is completely different. Based on the value of radius, PIL Gaussian blurring first construct two matrices $u$ and $v$ of size $1\times k$ and $k\times1$ where both $k$ and the scalar values in the matrix are determined by the radius. Then, it applies 3 rounds of row-wise convolution using $u$ and then three rounds of column-wise convolution using $v$. Moreover, after each round of convolution, it applies a particular rounding operation. It is also worth noting that the operator first pads the image on each side to ensure the size of blurred image is the same as the original. The padding is done using a special variation of reflection. All the operations are implemented in C. In order to perform the optimization, we implement the PIL blurring, with all the particularities, using pytorch functional operations so that we can backpropagate though and calculate gradients. Also, for the rounding operations, although they are not differentiable in theory, we implemented them as pytorch modules and defined their gradients to be always 1. Figure \ref{fig:PIL} shows our attack on PIL blurring.
    \item OpenCV: We also consider the OpenCV library and their blurring operation. This operation takes the dimensions of 2d kernel  and applies creates a uniform Kernel (which is same as Gaussian Kernel with standard deviation set to infinity infinity). Then after that there is a specific quantization step applied. OpenCV also uses reflection to pad the image according to the kernel size so that the dimensions of the blurred image remains same as the original. Similar to PIL, we implement these operation using differentiable modules in Pytorch and apply our attack.

    \item Masked Blurring: We also consider attacking masked blurring. We use the implementation of \cite{yang2022study} for face-blurring on ImageNet. Their approach uses PIL blurring to blur the entire image. Then it blurs the mask itself to make the blurring smoother. Finally it interpolates between the mask area 
    and then original image using the PIL Blend operation. To attack this, we need to implement a differentiable variant of blending operation. 
    
\end{itemize}

 \begin{figure}[h!]
    \centering
    \begin{minipage}{0.9\textwidth}
        \centering
        \includegraphics[width=\textwidth]{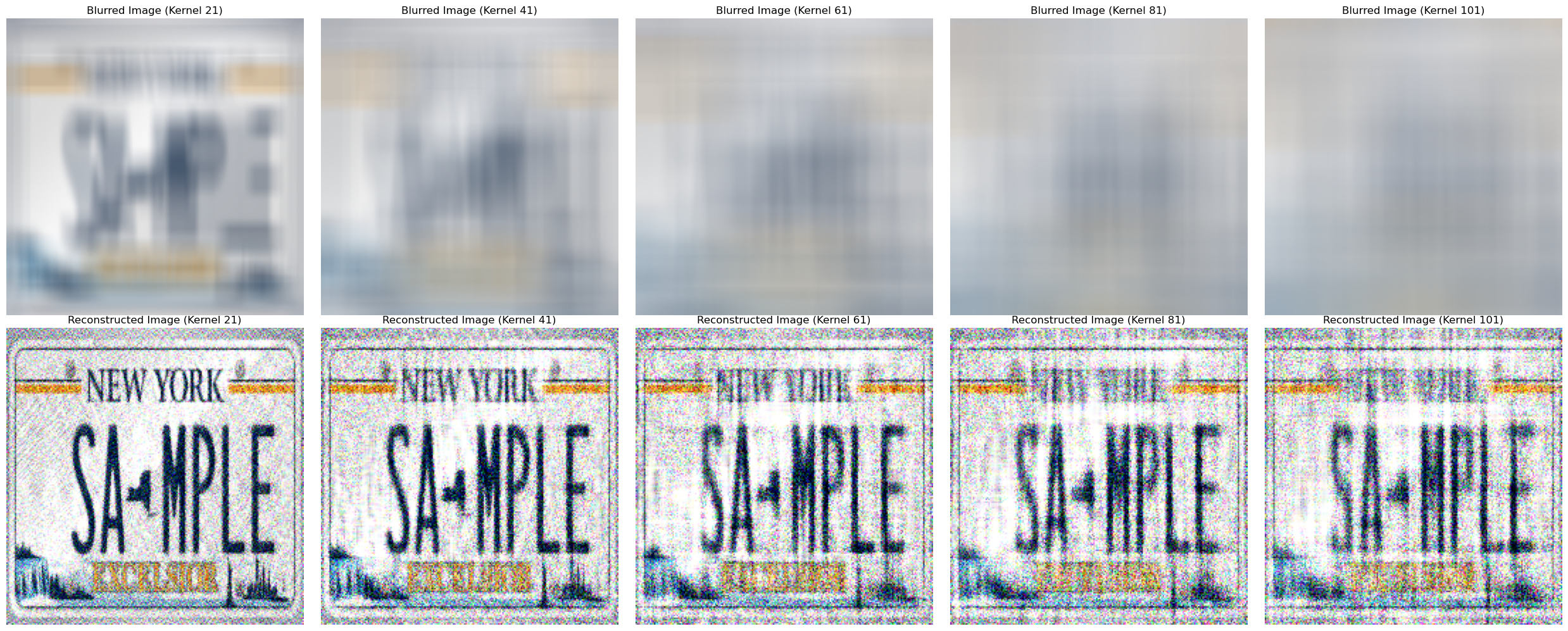}
    \end{minipage}\hfill
    
    \caption{We use an image of a license plate, resize it to 210x210 pixels and use OpenCV blurring with various kernel sizes. Then we use our attack to reconstruct the image. Our attack is initialized randomly (Gaussian noise) and uses learning rate 0.1 and momentum of.9. We run the attack for 5000 iterations.}
    \label{fig:OpenCV}

\end{figure}

\paragraph{Can we attack Pixelization?} Observe that the optimization~\ref{eqn:rev} is very general, and in theory, covers virtually any obfuscation operator $G$ including pixelization. However, we only expect the optimization to succeed in the sense that the solution $x^*$ is close to the original image $x$ only under certain conditions. For example, if $G$ involves considerable information loss, then the optimization problem will have many solutions, and the discovered solution $x^*$ will likely be quite far off from $x$. Pixelization is such transformation. Although Pixelization is also a linear transformation, its low-rank creates a significant information loss. For example, 
for pixelization, the pixelized image itself is a perfect solution to the optimization problem, simply because $G(G(x))=G(x)$ for any image $x$. The vast number of perfect solutions prevents the attack from reconstructing the original image.

%% file: theory_new.tex
\section{Analysis}

We next provide a brief intuitive analysis of the kind of privacy offered by the different obfuscation methods. For our analysis, we assume that the entire image consists of private information and is obfuscated; thus, an analysis of how side-information from background cues may impact privacy loss is out of scope for this section. 

\subsection{Privacy Framework and Definitions}

We will use $x \in \mathbb{R}^d$ to denote the unobfuscated image. The most popular way to provide a rigorous privacy guarantee on obfuscation mechanisms is to use differential privacy~\cite{dwork2006calibrating}; however, a differentially private obfuscation mechanism would requires that the obfuscated version of any two images -- for example, any two faces belonging to any two people -- are indistinguishable. This is a very stringent condition that is highly challenging to achieve for data release, and is perhaps a little more stringent than what is required for obfuscating faces and license plates so that they cannot be identified. 

Instead, we consider as our privacy framework a (slight variation of) the {\em{profile-based privacy}}~\cite{geumlek2019profile} framework, which does not have such a stringent requirement. Specifically, suppose there are $K$ profiles or identities, and data from the $k$-th profile is drawn from a distribution $P_k$. For example, if the $k$-th profile represents person $k$, then $P_k$ is the distribution formed by all images of his face taken from different angles and different lighting conditions. We say that the profiles $k$ and $l$ are indistinguishable if $P_k$ and $P_l$ overlap considerably. More formally, 

\begin{definition}[Indistinguishability] 
We say that profile $k$ is $(\alpha, \epsilon)$-indistinguishable from profile $l$ if the corresponding distributions $P_k$ and $P_l$ satisfy:
\[ \max(D_{\alpha} ( P_k, P_l), D_{\alpha} (P_l, P_k)) \leq \epsilon, \]
where $D_{\alpha}$ is the $\alpha$-Renyi divergence.
\end{definition}

Observe that $\epsilon$ and $\alpha$ are privacy parameters where lower $\epsilon$ and higher $\alpha$ imply more privacy. In general, a profile will differ from others to different degrees. Going back to our example of images of people, if $a$ and $b$ are siblings who look similar, then the distributions $P_a$ and $P_b$ will overlap quite a bit; in contrast, if $a$ and $b$ are people who look very dissimilar, then $P_a$ and $P_b$ will not overlap as much. 

Let $M$ be an obfuscation mechanism, randomized or not. For a profile $k$, we use the notation $M(P_k)$ to denote the distribution induced by applying $M$ to instances drawn from the distribution $P_k$. The main idea behind our privacy analysis is as follows. Usually, any two profiles $k$ and $l$ are quite distinct and the corresponding distributions $P_k$ and $P_l$ do not overlap as much: in our example of images of people, we can usually tell people apart based on the images of their faces. However, an obfuscation mechanism $M$ offers privacy if {\em{after applying $M$}}, the corresponding distributions $M(P_k)$ and $M(P_l)$ overlap to a higher degree -- which means that people's images become indistinguishable after obfuscation.

Specifically, the kind of privacy we aim for is as follows. 

\begin{definition}[$(N, \alpha, \epsilon)$-Profile-Based Privacy]
We say that a mechanism $M$ operating on a group $G$ of profiles satisfy $(N, \epsilon)$-profile-based privacy if for every profile $k \in G$, we have a set of profiles $U_k$ of size $\geq N$ such that for all $k' \in U_k$, $P_k$ and $P_{k'}$ are $(\alpha, \epsilon)$-distinguishable.  
\end{definition} 

If each profile is a person, then this means that after applying $M$, images of every person will be indistinguishable from that of at least $N$ other people. Higher the value of $N$, higher is the amount of privacy offered. 


\subsection{Privacy Properties of Obfuscation Mechanisms}

Unlike differential privacy, the actual profile-based privacy guarantee offered by an obfuscation mechanism is highly data-dependent: $N$ and $\epsilon$ both depend on the distributions of the images, the people in our set, among other factors. Consequently, in contrast with differential privacy, it is impossible to prove a general privacy guarantee on a particular obfuscation mechanism that applies to all data distributions. 

Instead, we will investigate what happens to the indistinguishability of two specific profiles under different obfuscation mechanisms. Specifically, given two profiles $k$ and $l$, we investigate properties of the two distributions $M(P_k)$ and $M(P_l)$. If these distributions overlap significantly more than $P_k$ and $P_l$, then we say that $M$ offers good privacy. 

\paragraph{Gaussian Blurring.} We begin with Gaussian blurring. Observe that full-precision Gaussian blurring is a bijective transformation; therefore, we can apply Proposition~\ref{prop:gaussianblur} to conclude that it does not change the Renyi divergence between the distributions of the profiles -- and the indistinguishability parameter $\epsilon$ remains the same before and after Gaussian blurring. This corroborates our empirical observation that it is easily reversible.

\begin{proposition}\label{prop:gaussianblur}
Let $M_g$ be any bijective transformation. If $P_k$ and $P_l$ are any two distributions, then for any $\alpha > 0$,
\[ D_{\alpha} (M_g(P_k), M_g(P_l)) = D_{\alpha} (P_k, P_l).\]
\end{proposition}

\paragraph{Cropping.} Another easy case is cropping, where we replace the private information by a constant. Cropping collapses all profiles into a single (one-point) distribution and hence makes all pairs of profiles completely indistinguishable. This again corroborates our observation that it is the most private obfuscation mechanism. 

\begin{proposition} \label{prop:crop}
Let $M_{c}$ be the cropping transformation that replaces the private information by a constant. For any two profiles $k$ and $l$, and any $\alpha > 0$, $D_{\alpha}(M(P_k), M(P_l)) = 0$.
\end{proposition}

\paragraph{Pixelization.} A slightly more complex case is pixelization, where the private portion of the image loses information. In this case, we can show that the Renyi divergence between two profiles either remains the same or strictly decreases after pixelization.

\begin{proposition}\label{prop:pixelize}
Let $M_p$ be the pixelization transformation that pixelizes an image to its $b \times b$ version. Then, for any two profiles $k$ and $l$, and any $\alpha > 0$, $D_{\alpha}(M(P_k), M(P_l)) \leq D_{\alpha}(P_k, P_l)$.
\end{proposition}

Whether the Renyi divergence strictly decreases or remains the same depends on the exact distribution. A pathological example where it remains the same is when $P_k = \bf{1}(0)$ and $P_l=\bf{1}(1)$ are delta distributions at the all-zeros and all-ones image respectively. Here, pixelization will lose information, but will not make these images indistinguishable. In contrast, if $P_k = \mathcal{N}(x_k, \sigma^2)$ and $P_l = \mathcal{N}(x_l, \sigma^2)$ are overlapping Gaussians, then pixelization will increase their overlap. We expect standard settings to more similar to the second case than the first, and this is why pixelation appears to help in practice. 

\paragraph{DP-Pix.} Finally, the outcomes for DP-Pix is yet again similar to pixelization: it will either reduce or keep the Renyi divergence between profiles the same, depending on the amount of noise added and the exact distributions. However, for DP-Pix with spherical Gaussian noise with standard deviation $\sigma$, we can prove a slightly tighter bound as shown in the following proposition.

\begin{proposition}\label{prop:dppix}
 Let $M_d$ be the DP-Pix transformation where $N(0, \sigma^2)$ noise is added to each image block after pixelization. Then, 
 \[ D_{\alpha}(M_d(P_k), M_d(P_l)) \leq \frac{\alpha W_2(P_k, P_l)^2}{\sigma^2},\]
 where $W_2(P_k, P_l)$ is the $L_2$-Wasserstein distance between $P_k$ and $P_l$.
\end{proposition}

In summary, these propositions corroborate our observations -- Gaussian blurring is the least private due to retaining information, cropping the most private; pixelization and DP-pix are somewhere in between, depending on the resolution of pixelization, the radius of the Gaussian noise, and the data distribution.

%% file: experiments.tex
\section{Experiments}

\subsection{Privacy}

We evaluate privacy through two separate attacks -- reversal and discrimination -- the former being the stronger attack. Ideally, our experiments should measure the ability to recognize subjects from their obfuscated images, but doing so presents ethical concerns. To avoid violating the privacy of real people, we instead carry out our privacy evaluations on OCR data: publicly available data on handwritten digits, house numbers and license plates. 

\subsubsection{Reversal attacks}
\label{sec:reversal_exp}
In a reversal attack, the goal is to evaluate how well the private information in an image can be reconstructed from its obfuscated version. To this end, we first apply the obfuscation method followed by our reversal technique in Section~\ref{sec:reversal}; we then evaluate reversal quality by checking to see if a model trained on the clean images can ``read'' the reconstructed image. 

\paragraph{Datasets and Baselines} We apply our experiments on three datasets --  MNIST~\cite{lecun1998gradient} and SVHN~\cite{netzer2011reading}, and a dataset of license plate images~\cite{licenseplate}. MNIST is a simple and clean dataset of handwritten digits, while SVHN consists of images of digits obtained from house numbers on Google street view. For MNIST classification, we train a CNN with accuracy $98.8\%$ on the clean MNIST test set. For SVHN, we use a pre-trained classifier available at~\cite{svhn-ocr} which has $94.49\%$ accuracy on the clean test set. For the license plate dataset, we use a pre-trained OCR model that has been pre-trained for reading license plates~\cite{hassani2021escaping} and evaluate a random subset of $400$ samples. 

For all three datasets, we consider three obfuscation methods -- Gaussian blurring, pixelization and DP-Pix. Gaussian blurring is carried out using the PIL library. For MNIST and SVHN, we report results with two different values of the radius -- Small ($1/10$ times the length of the maximum diagonal, the setting used in~\cite{yang2022study}) and Large ($1/7$ times the length of the maximum diagonal). The license plate dataset consists of rectangular images of full license plates, instead of single digits or letters. In this case, we also use Gaussian blurring with two radii -- Large ($1/40$ times the maximum diagonal) and Small ($1/50$ times the maximum diagonal). For all three datasets, we consider $4 \times 4$, $2 \times 2$ and $1 \times 1$ pixelization, and DP-Pix with the same pixelization configurations followed by Gaussian noise with standard deviation $0.04$. 

In all cases, we report the classification accuracy before and after reversal. For the license plates dataset, we report the character-wise recognition accuracy using a standard pre-trained OCR license plate reader.

\begin{table}[t]
\centering
\begin{tabular}{l|l| c|c}
\toprule
\centering
Dataset & Method & Accuracy before Reversal & Accuracy after Reversal \\
\midrule
\multirow{ 8}{*}{SVHN} & Gaussian Blur (Small) &  56.77   &  84.22 \\
 & Gaussian Blur (Large) &  29.57  & 55.41 \\
 & Pixelization (4 x 4) &  25.87    &  25.88 \\
 & Pixelization (2 x 2) & 19.58 & 19.59 \\
 & Pixelization (1 x 1) & 19.55  & 19.56\\
 & DP-Pix (4 x 4) & 21.14  & 25.94 \\
 & DP-Pix (2 x 2) & 19.34 & 19.57 \\
 & DP-Pix (1 x 1) & 19.55 & 19.57 \\
\midrule
\multirow{ 8}{*}{MNIST} & Gaussian Blur (Small) & 47.8    &  79.82 \\
 & Gaussian Blur (Large) & 12.94   &  58.82 \\
 & Pixelization (4 x 4) &   23.5   &   23.43\\
  & Pixelization (2 x 2) & 14.7 & 14.71 \\
 & Pixelization (1 x 1) & 5.61 & 5.58 \\
 & DP-Pix (4 x 4) &  28.3 &  23.84 \\
 & DP-Pix (2 x 2) & 10.52 & 12.44 \\
 & DP-Pix (1 x 1) & 9.33 & 4.33 \\
\bottomrule
\end{tabular}
\centering
\caption{Results of the Reversal Attack on MNIST and SVHN\label{tab:reversal}}
\end{table}

\begin{table}[]
\centering
\begin{tabular}{l| c|c}
\toprule
\centering
Method & Accuracy before Reversal & Accuracy after Reversal \\
\midrule
 Gaussian Blur (Small) &  35   &  88 \\
 Gaussian Blur (Large) &  26  & 60 \\
 Pixelization (4 x 4) &  24.6    &  24.6 \\
 Pixelization (2 x 2) & 23.00 & 23.00 \\
 Pixelization (1 x 1) & 22.00  & 22.00\\
 DP-Pix (4 x 4) & 24.5  & 24.5 \\
DP-Pix (2 x 2) & 22.70 & 23.00 \\
 DP-Pix (1 x 1) & 21.90 & 22.00 \\
\bottomrule
\end{tabular}
\centering
\caption{Results of the Reversal Attack on License Plates\label{tab:reversallicense}}
\end{table}

\begin{table}[t]
\centering
\begin{tabular}{l|l| c}
\toprule
Dataset & Method & Attack Accuracy\\
\midrule
\multirow{ 4}{*}{MNIST} & Gaussian Blur (Small) &  97.58  \\
 & Gaussian Blur (Large) & 95.75  \\
 & Pixelization (4 x 4) &  83.54 \\
 & Pixelization (2 x 2) & 51.16 \\
 & Pixelization (1 x 1) & 22.06 \\
 & DP-Pix (4 x 4) &  80.87 \\
 & DP-Pix (2 x 2) &   47.73\\
 & DP-Pix (1 x 1) &  21.98 \\
\midrule
\multirow{ 4}{*}{SVHN} & Gaussian Blur (Small) & 84  \\
 & Gaussian Blur (Large) & 72  \\
 & Pixelization (4 x 4) &  55.07 \\
 & Pixelization (2 x 2) & 23.82\\
 & Pixelization (1 x 1) & 19.58 \\
 & DP-Pix (4 x 4) & 49.23  \\
 & DP-Pix (2 x 2) &   12.91\\
 & DP-Pix (1 x 1) &   15.42 \\
\bottomrule
\end{tabular}

\caption{Results of the Discrimination Attack\label{tab:discrimination}}
\end{table}

\paragraph{Results.} The results are summarized in Table~\ref{tab:reversal}; for baseline comparison, we also report the classification accuracy before the reversal attack. 

We see classification accuracy after reversal is relatively high for Gaussian blurring, but considerably lower for pixelization and DP-pix -- even at a fairly high degree of pixelization ($4 \times 4$). In addition, the accuracy is higher for smaller radii than larger.  This implies that the reversal attack is quite successful on Gaussian blurring, and that the attack success rate decreases with the radius of the blurring kernel. 

In contrast, the reversal attack is considerably less successful for DP-Pix and pixelization:  in fact, for $2 \times 2$ and $1 \times 1$ pixelization and DP-Pix, the results are close to chance. This corroborates our theoretical findings. An interesting side-observation is that SVHN digits already have relatively high accuracy before reversal -- this might be because some of the SVHN digits are slightly blurred already, while the MNIST digits are much less so. Another interesting side-observation is that the baseline accuracy of the license plate reader is close to $22\%$; this may be because the reader is trained on European license plates that have a fixed structure. 

\subsubsection{Discrimination attacks}

In a discrimination attack, our goal is to evaluate how well private information in an image can be discerned out of a few possible options, given an auxiliary sample of similarly obfuscated data. Observe that this is a much stronger and less realistic attack than the reversal attack -- it assumes that the adversary has access to a dataset of similarly obfuscated images and their true labels, which may not hold for images in the wild. 

To carry out this attack, we train a classifier on the obfuscated data with the correct labels, and then use it to ``read'' obfuscated images. We again use the two datasets MNIST and SVHN and the same experiment setup as in Section~\ref{sec:reversal_exp}. Observe that since we do not have enough labeled digits from the license plate dataset, the attack is not possible in this setting. 

Table~\ref{tab:discrimination} shows the results. We see again that the attacks are more effective on Gaussian blurring, and significantly less effective on pixelization and DP-Pix. The efficacy of the attacks decreases with decreasing granularity of pixelization; interestingly, there is not much difference between pixelization and DP-Pix, perhaps because of the low standard deviation of the added noise. 

Another observation is that the discrimination attack works reasonably well on MNIST, which corroborates the results of~\cite{mcpherson2016defeating}. However, they are considerably worse on SVHN. We suspect this might be because MNIST is a very clean dataset where the digits are easily teased apart, even when obfuscated, while SVHN is more realistic. This suggests that while pixelization and DP-Pix may allow for discrimination attacks on toy settings such as MNIST, they are considerably more private for more realistic data such as SVHN.

\subsection{Utility}

\begin{table}[t]
\centering
\begin{minipage}[t]{0.39\linewidth}
\centering
\resizebox{\linewidth}{!}{%
\begin{tabular}{l|cc}
\toprule
\textbf{Mitigation} & \multicolumn{2}{c}{\textbf{ImageNet Accuracy}} \\
~ & Top-1 & Top-5 \\
\midrule
None & 75.85 & 92.76 \\
\midrule
Gaussian blur & 74.98 {\tiny ({\color{red} -0.87})} & 92.36 {\tiny ({\color{red} -0.40})} \\
Pixelization ($1 \times 1$) & 75.36 {\tiny ({\color{red} -0.49})} & 92.45 {\tiny ({\color{red} -0.31})} \\
DP-Pix ($4 \times 4, \sigma=0.04$) & 75.33 {\tiny ({\color{red} -0.52})} & 92.53 {\tiny ({\color{red} -0.23})} \\
\midrule
Crop & 75.09 {\tiny ({\color{red} -0.76})} & 92.31 {\tiny ({\color{red} -0.45})} \\
\bottomrule
\end{tabular}
}
\caption{ImageNet classification \label{tab:imagenet}}
\end{minipage}
\begin{minipage}[t]{0.6\linewidth}
\centering
\resizebox{\linewidth}{!}{%
\begin{tabular}{l|cccc}
\toprule
\textbf{Mitigation} & \multicolumn{4}{c}{\textbf{Zero-shot Top-1 Accuracy}} \\
~ & ImageNet & CIFAR-10 & CIFAR-100 & GTSRB \\
\midrule
None & 64.83 & 92.10 & 72.75 & 35.52 \\
\midrule
Gaussian blur & 64.35 {\tiny ({\color{red} -0.48})} & 90.97 {\tiny ({\color{red} -1.13})} & 70.64 {\tiny ({\color{red} -2.11})} & 38.98 {\tiny ({\color{blue} +3.46})} \\
Pixelization ($1 \times 1$) & 64.83 {\tiny (-0.00)} & 92.90 {\tiny ({\color{blue} +0.80})} & 72.65 {\tiny ({\color{red} -0.10})} & 35.50 {\tiny ({\color{red} -0.02})} \\
DP-Pix ($4 \times 4, \sigma=0.04$) & 64.67 {\tiny ({\color{red} -0.16})} & 92.99 {\tiny ({\color{blue} +0.89})} & 71.76 {\tiny ({\color{red} -0.99})} & 37.30 {\tiny ({\color{blue} +1.78})} \\
\midrule
Crop & 64.38 {\tiny ({\color{red} -0.45})} & 91.74 {\tiny ({\color{red} -0.36})} & 71.95 {\tiny ({\color{red} -0.80})} & 32.94 {\tiny ({\color{red} -2.58})} \\
\bottomrule
\end{tabular}
}
\caption{MetaCLIP zero-shot classification \label{tab:metaclip}}
\end{minipage}

\end{table}

\begin{table}[t]
\centering

\begin{minipage}[t]{0.425\linewidth}
\centering
\resizebox{\linewidth}{!}{%
\begin{tabular}{l|cc}
\toprule
\textbf{Mitigation} & \multicolumn{2}{c}{\textbf{MS-COCO 2017}} \\
~ & \multicolumn{2}{c}{\textbf{Semantic Segmentation}} \\
~ & Box AP & Mask AP \\
\midrule
None & 38.6 & 35.2 \\
\midrule
Gaussian blur & 37.7 {\tiny ({\color{red} -0.9})} & 34.5 {\tiny ({\color{red} -0.7})} \\
Pixelization ($1 \times 1$) & 38.0 {\tiny ({\color{red} -0.6})} & 34.6 {\tiny ({\color{red} -0.6})} \\
DP-Pix ($4 \times 4, \sigma=0.04$) & 37.8 {\tiny ({\color{red} -0.8})} & 34.6 {\tiny ({\color{red} -0.6})} \\
\midrule
Crop & 37.7 {\tiny ({\color{red} -0.9})} & 34.4 {\tiny ({\color{red} -0.8})} \\
\bottomrule
\end{tabular}
}
\caption{Detectron2 MS-COCO 2017 semantic segmentation \label{tab:semantic_seg}}
\end{minipage}
\begin{minipage}[t]{0.565\linewidth}
\centering
\resizebox{\linewidth}{!}{%
\begin{tabular}{l|cccc}
\toprule
\textbf{Mitigation} & \multicolumn{4}{c}{\textbf{MS-COCO 2014}} \\
~ & \multicolumn{4}{c}{\textbf{Dense Pose Estimation}} \\
~ & Box AP & Mask AP & GPS AP & GPSm AP \\
\midrule
None & 61.0 & 66.9 & 63.5 & 65.2 \\
\midrule
Gaussian blur & 59.9 {\tiny ({\color{red} -1.1})} & 65.7 {\tiny ({\color{red} -1.2})} & 61.7 {\tiny ({\color{red} -1.8})} & 63.4 {\tiny ({\color{red} -1.8})} \\
Pixelization ($1 \times 1$) & 59.9 {\tiny ({\color{red} -1.1})} & 65.3 {\tiny ({\color{red} -1.6})} & 61.2 {\tiny ({\color{red} -2.3})} & 62.8 {\tiny ({\color{red} -2.4})} \\
DP-Pix ($4 \times 4, \sigma=0.04$) & 59.9 {\tiny ({\color{red} -1.1})} & 65.6 {\tiny ({\color{red} -1.3})} & 61.4 {\tiny ({\color{red} -2.1})} & 63.0 {\tiny ({\color{red} -2.2})} \\
\midrule
Crop & 59.8 {\tiny ({\color{red} -1.2})} & 65.0 {\tiny ({\color{red} -1.9})} & 61.0 {\tiny ({\color{red} -2.5})} & 62.8 {\tiny ({\color{red} -2.4})} \\
\bottomrule
\end{tabular}
}
\caption{Detectron2 MS-COCO 2014 dense pose estimation \label{tab:dense_pose}}
\end{minipage}

\end{table}

In this section, we evaluate the effect of different blurring methods on model utility. We consider four different tasks: image classification on ImageNet, vision-language modeling with MetaCLIP, semantic segmentation, and dense pose estimation.

\paragraph{Mitigations.} For this experiment, we focus on obfuscating faces in image datasets. We first use RetinaFace \cite{deng2020retinaface, serengil2020lightface} to detect the face bounding box, and then apply one of the following mitigations to obfuscate the face region (see section \ref{sec:obfuscation} for details):
\begin{itemize}
    \item \emph{None}: Original unaltered image with no mitigation applied. This serves as an upper bound on utility.
    \item \emph{Gaussian blur}: The Gaussian blur algorithm implemented in \cite{yang2022study}.
    \item \emph{Pixelization}: We consider $1 \times 1$ pixelization across the face bounding box. In other words, the face region is replaced by the average value inside the face bounding box.
    \item \emph{DP-Pix}: We consider DP-Pix with $4 \times 4$ pixelization and add $\sigma=0.04$ Gaussian noise to the pixelized face image.
    \item \emph{Crop}: We replace the face region with a black square to remove all information in the region. In principle, this serves as a lower bound on model utility that any non-trivial mitigation should exceed.
\end{itemize}

\paragraph{ImageNet classification.} We train a ResNet-50~\cite{he2016deep} model on the ImageNet-1K dataset with different face-blurring mitigations applied. Result is shown in Table \ref{tab:imagenet}. The top row shows top-1 and top-5 accuracy without any mitigation, which serves as an upper bound. In comparison, all mitigations achieve slightly lower but comparable accuracy.

\paragraph{Vision-language modeling with MetaCLIP.} Vision-language models (VLMs) such as CLIP \cite{radford2021learning} serve as fundamental building blocks in modern AI. These models are typically trained on billion-scale web datasets that contain image-text pairs, and can learn to associate images with corresponding captions in a way that is transferable to different downstream tasks.

To evaluate the utility effect of mitigations in this setting, we train a MetaCLIP \cite{xu2023demystifying} model from scratch on its 2.1B training dataset with different mitigations applied. We follow the same hyperparameter setting as in the original paper, utilizing a ViT-B-32 backbone and processing through a total of 12.8B (non-unique) image-text pairs. To evaluate model utility, we measure the model's zero-shot classification accuracy on four image datasets (without any mitigation): ImageNet-1K, CIFAR-10, CIFAR-100 and GTSRB.
Table \ref{tab:metaclip} shows zero-shot top-1 accuracy on the four datasets. Similar to Table \ref{tab:imagenet}, we observe a slight degradation of performance when mitigations are applied but the effect is negligible. All mitigations achieve similar performance.

\paragraph{Semantic segmentation and dense pose estimation.} The above two evaluations consider classification tasks that do not require any information around the face region. This is evident by the small performance gap between the top and bottom rows, \emph{i.e.}, under no mitigation vs. having the entire face region cropped. To stress test the mitigations on a more difficult task, we evaluate on MS-COCO semantic segmentation and dense pose estimation. \emph{Semantic segmentation} is a classic computer vision task where the objective is to localize different objects in an image with a bounding box or pixel-wise mask. \emph{Dense pose estimation} aims at identifying a correspondence between pixels in an image and object geometry. Both tasks are more sensitive to image modifications since more detailed information is required to localize the correct object or estimate the pose of a human/animal.

We evaluate different mitigations by training ResNet-50 backbone models for semantic segmentation and dense pose estimation using the Detectron2\footnote{\url{https://github.com/facebookresearch/detectron2/tree/main}} framework. We use Mask R-CNN \cite{he2017mask} for semantic segmentation and Continuous Surface Embedding \cite{neverova2020continuous} for dense pose estimation. The performance metric is \emph{average precision} (AP) of the prediction; the higher the better. We report the relevant metrics for semantic segmentation in Table \ref{tab:semantic_seg} and for dense pose estimation in Table \ref{tab:dense_pose}. In both cases, there is a more noticeable performance drop for all mitigations. Importantly, Pixelization and DP-Pix achieve similar or better performance compared to Gaussian blurring. This evaluation demonstrates that Pixelization and DP-Pix can be good alternatives to Gaussian blur utility-wise.